\documentclass[letterpaper, 10 pt, conference]{ieeeconf}  

\IEEEoverridecommandlockouts                              

\usepackage{amsmath} 
\usepackage{graphicx}
\title{\LARGE \bf
Grip as Needed, Glide on Demand: \\Ultrasonic Lubrication for Robotic Locomotion
}

\author{Mostafa A. Atalla$^{1,2,\ast}$, Jack Cumming$^{1}$, Daan van Bemmel$^{1}$,\\Paul Breedveld$^{1}$, Micha\"el Wiertlewski$^{2,\dagger}$, and Aim\'ee Sakes$^{1,\dagger}$%
\thanks{$^{\dagger}$These authors share last authorship}
\thanks{$^{1}$M. A. Atalla, D. van Bemmel, J. Cummings, P. Breedveld, and A. Sakes are with the Department of BioMechanical Engineering, Delft University of Technology (TU Delft), 2628~CD Delft, The Netherlands.}%
\thanks{$^{2}$M. A. Atalla and M. Wiertlewski are with the Department of Cognitive Robotics, Delft University of Technology (TU Delft), 2628~CD Delft, The Netherlands.}%
\thanks{$^{\ast}$Corresponding author: {\tt\small m.a.a.atalla@tudelft.nl}}
}

\begin{document}

\maketitle
\thispagestyle{empty}
\pagestyle{empty}

\begin{abstract}
Friction is the essential mediator of terrestrial locomotion, yet in robotic systems it is almost always treated as a passive property fixed by surface materials and conditions. Here, we introduce ultrasonic lubrication as a method to actively control friction in robotic locomotion. By exciting resonant structures at ultrasonic frequencies, contact interfaces can dynamically switch between "grip" and "glide" states, enabling locomotion. We developed two friction control modules: a cylindrical design for lumen-like environments and a flat-plate design for external surfaces, and integrated them into bio-inspired systems modeled after inchworm and wasp ovipositor locomotion. Both systems achieved bidirectional locomotion with nearly perfect locomotion efficiencies that exceeded $90\%$. Friction characterization experiments further demonstrated substantial friction reduction across various surfaces, including rigid, soft, granular, and biological tissue interfaces, under dry and wet conditions, and on surfaces with different levels of roughness, confirming the versatility of ultrasonic lubrication for locomotion applications. These findings establish ultrasonic lubrication as a viable active friction control mechanism for robotic locomotion, with the potential to reduce design complexity and improve the efficiency of robotic locomotion systems.
\end{abstract}

\section{Introduction}
\begin{figure}
    \centering
    \includegraphics[width=\linewidth]{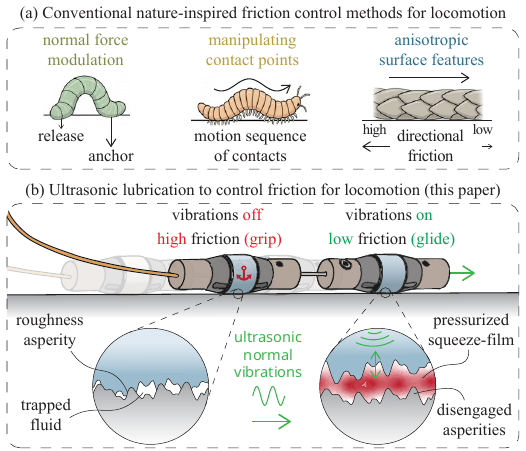}    \caption{\textbf{Concept of ultrasonic lubrication for locomotion.} (a) Conventional nature-inspired friction control methods for locomotion through normal force modulation, contact sequencing, or anisotropic surface features. (b) The proposed approach uses transverse ultrasonic vibrations to actively control friction at the contact interface. These vibrations pressurize the trapped fluid between the microscopic roughness asperity junctions, creating a pressurized fluid cushion that disengages the two surfaces, allowing for controllable switching between high-friction `grip' (vibrations off) and low-friction `glide' (vibrations on) states to drive locomotion.}
    \label{fig:UltrasonicLubricationLocomotionConcept}
\end{figure}

Biological organisms achieve remarkable locomotion by exploiting frictional interactions at the body–environment interface. Inchworms advance through sequential frictional anchoring of body segments combined with extension–contraction cycles, while earthworms propel through retrograde peristaltic waves aided by ventral setae to modulate frictional grip along the body~\cite{Saga2016,Xu2022,Calisti2017,Das2023}. Parasitic wasps steer ultra-thin ovipositors through dense substrates using reciprocating interlocking valves (sliders) that create frictional anisotropy, enabling substrate penetration without buckling~\cite{van2020,Sakes2020,Nikelshparg2023}, and limbless organisms, such as snakes, utilize overlapping ventral scales to create directional frictional asymmetry that enables them to propel. Fundamentally, these biological systems control friction at the contact interface to achieve locomotion through three primary strategies: by modulating the normal force at the contact points to anchor or release, by coordinating the motion sequence of those contact points to generate friction differential, or by employing anisotropic surface features to create directional friction asymmetry, as illustrated in Fig.\ref{fig:UltrasonicLubricationLocomotionConcept}(a).

These locomotion strategies have inspired a wide range of robots for confined and challenging environments, from peristaltic soft crawlers for search-and-rescue in rubble~\cite{Xu2022,Blewitt2024,Kulkarni2025}, to earthworm-inspired robots for pipeline inspection~\cite{joey2019,Elankavi2022,sebastian2025}, to inchworm-inspired robots for medical applications such as self-propelling endoscopic capsules~\cite{Manfredi2019,ciuti2016,Ahmed2024}. Across these developments, robotic systems have largely sought to replicate the friction-control principles observed in biological systems. To emulate normal-force modulation, inflatable fluidic structures have been used to vary contact pressure and thereby control the anchoring and release of robot segments~\cite{menciassi2003, Manfredi2019}. To reproduce contact sequencing, cam-driven mechanisms and fluidic control systems have been employed to coordinate the motion of multiple sliders and generate the desired friction differential~\cite{Sakes2020,Bloemberg2025WaspBiopsy,Atalla2026}. In parallel, directional pads with scale-like features have been introduced to create directional frictional asymmetry~\cite{Saab2019,Manoonpong2016,Calisti2017,Tramsen2018,Kim2023}.

However, these friction-control techniques impose inherent limitations in the design of locomotion systems. Strategies based on modulating normal force require the robot to locally expand or apply elevated contact forces to anchor, which can limit adaptability in fragile or tightly confined environments and often require dedicated actuators. Strategies based on coordinating the motion sequence of multiple contacts demand precise gait control through dedicated actuators and/or sophisticated motion-transmission systems and often require a large number of contacts to generate sufficient friction differential for traction, significantly increasing the overall system complexity. Strategies based on anisotropic surface features rely either on fixed microscopic textures that cannot adapt to changing locomotion demands, or on controllable macroscopic scales that further complicate the design by requiring onboard mechanisms for their deployment and actuation. These limitations suggest the need for an alternative friction-control method that can regulate the frictional properties of the interface independently of normal load, number of contacts and surface texture. 

\begin{figure*}[t!]
    \centering
    \includegraphics[width=\linewidth]{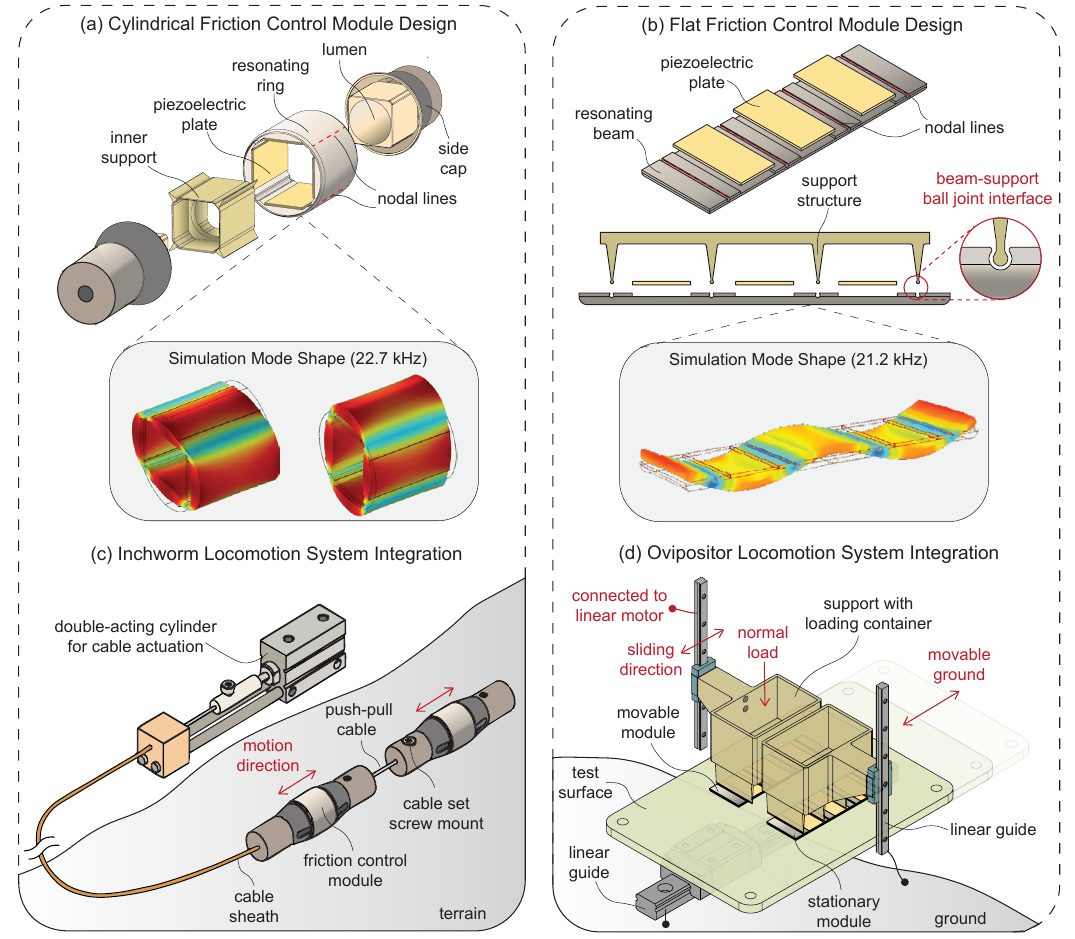}
    \caption{\textbf{Design and integration of cylindrical and flat friction control modules for bio-inspired locomotion.} (a) Cylindrical module: a $10~\mathrm{mm}$ ring-shaped resonator with four bonded piezoelectric plates, supported at nodal lines by an internal frame to minimize interference with oscillation. Finite element simulation predicted the second flexural resonance mode at $22.7~\mathrm{kHz}$. (b) Flat module: a $32 \times 10 \times 1~\mathrm{mm}$ slider resonator with piezoelectric plates bonded at antinodes and supported via a ball-joint groove at nodal lines. Simulation results predicted the third flexural resonance mode at $21.2~\mathrm{kHz}$. (c) Inchworm-inspired locomotion system: two cylindrical modules connected by a push–pull cable-sheath mechanism and actuated by a double-acting cylinder, enabling bidirectional motion through selective activation of each module. (d) Ovipositor-inspired locomotion system: two flat modules, one fixed and one mounted to a linear actuator, enabling bidirectional motion through selective activation of the movable module during the motion cycle.}
    \label{fig:design}
\end{figure*}
We present ultrasonic lubrication as an active friction control method for robotic locomotion. Unlike approaches that vary normal force or use surface anisotropy, ultrasonic lubrication directly modulates the coefficient of friction, independently of normal load, contact count, and surface design. This method uses transverse ultrasonic vibrations at the interface to generate a pressurized fluid film that lubricates the contact interface, thereby reducing the coefficient of friction for smooth low-friction sliding on demand, as illustrated in Fig.~\ref{fig:UltrasonicLubricationLocomotionConcept}(b). By dynamically adjusting the coefficient of friction at the body-environment interface, a locomotor can determine when and where to grip or glide, moving beyond the limits of passive interfaces and opening up new possibilities for more efficient and less mechanically complex robotic locomotion systems. To investigate the viability of this approach, we present two ultrasonic lubrication module designs, which we implement and integrate into two bio-inspired locomotion systems inspired by inchworm and ovipositor gaits, respectively, and evaluate their performance under different locomotion conditions.

In the remainder of this paper, Section~\ref{sec:principle} introduces the working principle of ultrasonic lubrication. Section~\ref{sec:design} presents the design, fabrication, and integration of two friction control modules. Section~\ref{sec:experiments} describes the experimental setup and protocols, and Section~\ref{sec:results} reports the vibration, locomotion, and friction modulation results. Finally, Section~\ref{sec:conclusion} summarizes the findings and discusses their implications for future robotic locomotion systems.


\section{Working Principle}\label{sec:principle}
\subsection{Ultrasonic Lubrication}
When two solid bodies come into contact and slide against each other, friction emerges due to the interaction of a discrete number of microscopic surface asperities, which form micro-junctions at the interface. The sum of the surface areas of these micro-junctions constitutes the real contact area, which is much smaller than the apparent area of contact. As a result, a large fraction of the apparent contact area does not engage directly, leaving fluid from the surrounding environment trapped at the interface (Fig.\ref{fig:UltrasonicLubricationLocomotionConcept}(b)) that form a thin interfacial layer, commonly described as a squeeze film.

Introducing transverse ultrasonic vibrations to the contact bodies causes the trapped fluid to undergo nonlinear compression, generating excess pressure within the fluid film. This positive pressure separates the two surfaces, disengaging the roughness asperities and lubricating the interface, thus reducing the coefficient of friction, a mechanism known as ultrasonic lubrication or ultrasonic friction modulation~\cite{Wiertlewski2016, Friesen2017}. The degree of separation between the two surfaces and therefore the lubrication effect is proportional to the vibration amplitude, allowing precise and active control of the coefficient of friction at the interface on demand~\cite{Wiertlewski2016}.

Ultrasonic lubrication has been successfully applied in a range of contexts where friction control is critical, including tactile displays \cite{Winfield2007}, non-contact object manipulation \cite{Gabai2019}, and squeeze-film bearings \cite{Zhao2013,Shi2019}. Furthermore, it has been shown to operate effectively in submerged liquid environments~\cite{Atalla2023, atalla2024}, highlighting its versatility across diverse operating conditions. This ability to switch between high- and low-friction states in different environments and conditions on demand suggests the potential of ultrasonic lubrication for applications requiring active friction control, such as robotic locomotion.

\subsection{Biological Locomotion Use Cases}
To investigate the potential of ultrasonic lubrication as an active friction control method for locomotion, we examined two representative biological gaits with complementary modes of locomotion. The first is inchworm locomotion, which operates through sequential anchoring and advancement of body segments. The second is peristaltic slider locomotion, exemplified by the wasp ovipositor, which relies on the reciprocating sliding of multiple valves. These two use cases were selected because they have inspired numerous robotic designs, making them well-suited as benchmarks for proof-of-concept demonstrations. 

The inchworm achieves locomotion through discrete phases of anchoring and advancing. It alternately grips the substrate with its front and rear prolegs, contracts its body to bring the rear forward, then extends again to project the front forward. At each stage of the cycle, one part of the body remains fixed while the other is moving, ensuring sequential progress. This stop-and-go strategy produces net displacement without requiring continuous wave propagation and has inspired a range of robotic implementations, often termed “inchworm robots” or “two-anchor crawlers,” where sequential anchoring is achieved with frictional pads, clamps, or suction elements.

Parasitic wasps achieve locomotion of their ovipositor through a peristaltic slider mechanism, where three slender valves—one dorsal and two ventral—are held together by a tongue-and-groove connection that constrains them radially while permitting longitudinal sliding. During penetration, locomotion is generated by sequential reciprocation: while one valve advances forward into the substrate, the other two remain stationary and provide anchoring through friction, counteracting the penetrating force. The advancing valve then becomes stationary while another valve moves forward, and this cycle repeats across all three valves. After each has advanced in turn, the valves reset in unison, producing a net forward displacement of the ovipositor. By iterating this cycle, the ovipositor is able to steadily penetrate deep into host tissue or other substrates with minimal overall insertion force.

\section{Design and Implementation }\label{sec:design} 
\subsection{Design Requirements}
To implement the two locomotion strategies using ultrasonic lubrication, the first step is to develop a friction control module, which is the fundamental building block of the locomotion system. We present two module designs: a cylindrical block intended for locomotion within lumens (internal locomotion) and a flat-plate block designed for locomotion on external surfaces. For ultrasonic lubrication to be effective, each module must be able to produce a vibration amplitude of $\geq20~\mathrm{kHz}$ at an ultrasonic frequency of $\geq20~\mathrm{kHz}$~\cite{Biet2007,Watanabe1995}.

Both modules share a common design principle. The resonating structure is engineered with a flexural resonance mode in the ultrasonic range. Piezoelectric plates are positioned at the antinodes of this mode so that, when activated, they excite the structure at its designed flexural resonance mode. To allow the structure to vibrate freely, structural supports are positioned at the nodal lines of the resonance mode, where the displacement is nearly zero. These supports, implemented through grooves and needle-like contact elements, allow free rotation at the nodes, and thus reduce interference with vibration. This approach enables the resonating structure to act as a mechanical amplifier, achieving the vibration amplitudes required for effective ultrasonic lubrication (Fig.~\ref{fig:design}(a)(b)).

\subsection{Cylindrical Friction Control Module Design}
The cylindrical module consists of a ring-shaped resonator with four bonded piezoelectric plates, designed to operate in its second flexural resonance mode, as illustrated in Fig.\ref{fig:design}(a). The prototype used for characterization had an outer diameter of $10~\mathrm{mm}$, following the design of~\cite{atalla2024}. The ring is supported internally by a frame with needle-like edges contacting the resonator at nodal lines. Finite element analysis, using stainless steel as the material, predicted a resonance frequency of $22.7~\mathrm{kHz}$ and a maximum vibration amplitude of approximately $4~\mu\mathrm{m}$ per $100~\mathrm{V}$ excitation.

\subsection{Flat Friction Control Module Design}
The flat module is a rectangular slider resonator ($32 \times 10 \times 1~\mathrm{mm}$) with three piezoelectric plates bonded at the antinodes, designed to operate in its third flexural mode, as shown in Fig.~\ref{fig:design}(b). The slider is supported at its nodal lines through small circular cutouts that accommodate a ball-joint groove and thin-legged supports, constraining the structure while allowing free rotation. Finite element analysis, using stainless steel as the material, predicted a resonance frequency of $21.2~\mathrm{kHz}$ and a maximum vibration amplitude of approximately $4.73~\mu\mathrm{m}$.

\begin{figure}
    \centering
    \includegraphics[width=\linewidth]{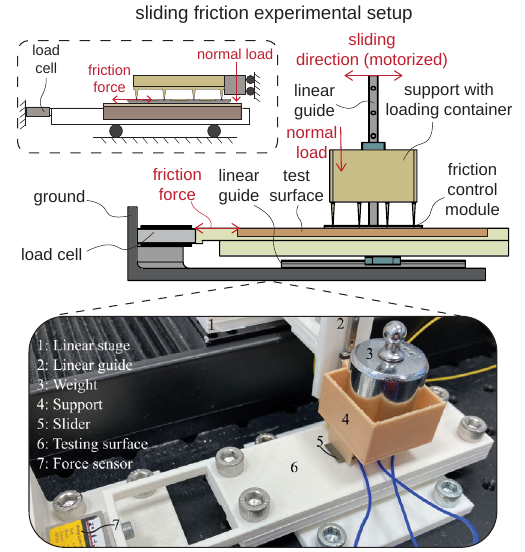}
    \caption{\textbf{Experimental setup for the friction characterization experiments.} A miniature load cell mounted along the sliding axis measured friction forces via a bracket guided on a horizontal stage, while the slider was mounted to a vertical linear guide for motorized horizontal motion and free vertical motion. A 100 g weight ($\approx$1 N) provided the normal load. Two protocols were tested: (i) sliding with and without continuous vibration at a voltage amplitude of 140~V to assess stability of the lubrication effect when activated, and (ii) sliding with the excitation voltage amplitude ramped up from 0 to 140 V to evaluate dynamic friction modulation. Tests were performed on dry and wet PLA, sandpapers of different grit sizes, granular soil, and ex-vivo porcine intestinal tissue.}
    \label{fig:expsetup}
\end{figure}

\subsection{Prototype Development}
Friction control modules, both flat and cylindrical, were fabricated from stainless steel 316 using electrical discharge machining (EDM). Piezoelectric plates (SM112, SteminC, FL, USA) were cut to size and bonded into pre-designed slots with a two-component, electrically insulating adhesive (3M Scotch-Weld Epoxy Structural Adhesive DP490). Each plate included two electrodes on one side, with the back electrode bonded to the resonator surface. Wires were soldered to the exposed electrodes and arranged so that consecutive plates were driven with a phase shift of $180^{\circ}$, ensuring excitation of the target resonance mode. The support structures were produced using a stereolithography (SLA) 3D printing (Formlabs, USA), providing the precision required for the fine support legs interfacing with the resonating structures.

\subsection{Locomotion System Integration}
The inchworm-inspired locomotion system was realized using two cylindrical friction control modules with an extensor element positioned between them, as illustrated in Fig.\ref{fig:design}(c). The extensor was implemented through a push–pull cable-sheath mechanism, in which the sheath was anchored to the first module while the inner cable passed through and was fixed to the second module. The push–pull assembly was driven by a double-acting linear cylinder that actuated the extension and contraction. A computer-programmed control system coordinated the actuation sequence by selectively activating each friction control module based on the desired direction of motion and the phase of the motion cycle.

The ovipositor-inspired locomotion system was constructed using two flat friction control modules, as illustrated in Fig.\ref{fig:design}(d). One module was fixed in place and the other was mounted to a linear actuator that provided reciprocating motion while the ground was mounted on a linear bearing to slide freely. A computer-programmed control system was then used to selectively switch the vibrations of the moving slider on or off, depending on the desired direction of motion and the phase of the motion cycle. By fixing one slider and actuating the other, while allowing the contacting ground surface to slide freely, the same principle of ovipositor locomotion could be demonstrated in a simpler and functionally equivalent manner that is better suited for experimental evaluation.

\section{Experimental Evaluation}\label{sec:experiments} 
To validate the concept of the friction control modules and characterize their performance, we conducted three experiments. The first experiment aimed to characterize and validate the vibration performance of individual modules. The second was a proof-of-concept experiment to demonstrate locomotion enabled by ultrasonic lubrication in both bio-inspired designs and to quantify the resulting locomotion efficiency. The third aimed to assess the ability of the modules to modulate friction across surfaces with different materials and conditions.

\subsection{Vibration Characterization Experiment}
The resonance properties of the friction control modules were evaluated by measuring the resonance frequency and the amplitude of the vibration through two complementary tests. First, a frequency response measurement was used to identify the resonant frequency. Second, a voltage sweep was performed at this frequency to quantify the vibration amplitude as a function of the input voltage. Both unsupported and supported sliders were tested to assess the influence of the support structure, and each experiment was repeated six times.  

The vibration response was measured using a laser Doppler vibrometer (Polytec OFV-5000 Vibrometer Controller). The Laser Doppler Vibrometer tracked the velocity of a marked anti-nodal point at the center of the slider, while the prototypes were powered by a piezo amplifier (PiezoDrive PD200X4 Voltage Amplifier). The Laser Doppler Vibrometer and amplifier were connected to a data acquisition system (National Instruments USB X Series Multifunction DAQ), which interfaced with a PC for control and data logging.  

For the frequency response test, the modules were activated, and the excitation frequency was swept from 20 to 24 kHz at maximum voltage to determine the resonant frequency. For the amplitude test, the frequency was fixed at the identified resonance frequency, and the input voltage was increased from 0 to 100 V in steps of 10 V to record the corresponding vibration amplitudes.  

\subsection{Proof-of-concept Locomotion Experiment}
We tested both locomotion systems on rigid 3D-printed PLA substrates to demonstrate the viability of ultrasonic lubrication for locomotion and to quantify locomotion efficiency. The inchworm-inspired system was evaluated on a semi-cylindrical substrate that matched the geometry of the cylindrical friction control modules, while the ovipositor-inspired system was tested on a flat substrate. Each system was first actuated without ultrasonic lubrication to establish a baseline, after which the same motion sequence was repeated with the ultrasonic lubrication activated. For each test, the distance traveled by the system $d_{actual}$ was measured and compared to the actuator stroke $d_{theoretical}$—the theoretical displacement the locomotion system should achieve without slipping—to determine the locomotion efficiency as follows:

\begin{equation}
    \eta=\frac{|d_{actual}-d_{theoretical}|}{d_{theoretical}} \times 100
\end{equation}

\subsection{Friction Characterization Experiment}
Following the proof-of-concept, we conducted sliding friction experiments using only the flat friction control module to evaluate the slider’s ability to modulate friction across a range of surface conditions: a rigid 3D-printed PLA surface under both dry and wet conditions, a granular soil surface, and surfaces with varying roughness using coarse (150 grit) and fine (240 grit) sandpapers. Finally, we performed tests on wet ex-vivo biological tissue (porcine intestinal tissue) to explore the viability for medical applications. 

Friction forces were measured with a miniature load cell (Futek LSB200) mounted along the sliding axis and connected to the test surface through a mounting bracket, which was supported by a horizontal linear guide (IKO Nippon Thompson LWLC3C1R60T0H, LWL) to allow free motion in the direction of the friction force, as illustrated in Fig.\ref{fig:expsetup}. The slider was attached to a linear actuator (Thorlabs NRT150/M) through a vertical linear guide, providing controlled horizontal displacement while allowing free vertical motion of the slider to apply the normal load using weights. 

\begin{figure}
    \centering
    \includegraphics[width=\linewidth]{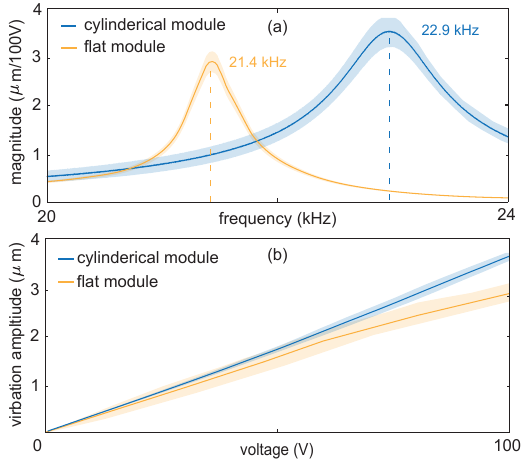}
    \caption{\textbf{Vibration characterization of cylindrical and flat friction control modules.} (a) Frequency response showing resonance peaks at $22.9~\mathrm{kHz}$ for the cylindrical module and $21.4~\mathrm{kHz}$ for the flat module. 
    (b) Vibration amplitude as a function of applied voltage, exhibiting linear growth up to an amplitude of $\approx4~\mu\mathrm{m}$ at 100~V. Both modules achieved vibration amplitudes exceeding $2~\mu\mathrm{m}$ at ultrasonic frequencies ($\geq20~\mathrm{kHz}$), fulfilling the design requirements of ultrasonic lubrication.}
    \label{fig:vibrometry}
\end{figure}
In this experiment, two sets of tests were conducted. In the first, the friction control module was loaded with a 100 g weight ($\approx$1 N normal load) and driven forward and backward over a 10 mm distance at 1~mm/s. The test was performed first without vibrations and then with continuous activation at a voltage amplitude of 140~V to evaluate the stability of the lubrication effect under steady excitation. In the second set of tests, the module’s ability to modulate friction was assessed by sliding it over a 5 mm distance at 1 mm/s under the same 1~N load while gradually ramping the input voltage amplitude from 0 to 140~V. To quantify the module's friction modulation capability, we used the friction reduction percentage, which was calculated as follows:
\begin{equation} \label{eq:reduction}
    \mathrm{friction\ reduction\ (\%)} = \frac{\mu_{off}-\mu_{on}}{\mu_{off}}
\end{equation}
where $\mu_{off}$ is the average coefficient of friction in the non-lubricated 'off' case, and $\mu_{on}$ is the coefficient of friction in the lubricated 'on' case.

\section{Results and Discussion}\label{sec:results}
The vibration characterization experiments revealed the resonance properties of both types of friction control modules. The cylindrical ring prototype resonated at $22.9~\mathrm{kHz}$, at which it produced a vibration amplitude of $\approx 3.7~\mu\mathrm{m}/100\mathrm{V}$. The flat module prototype resonated at $21.4~\mathrm{kHz}$ (Fig.~\ref{fig:vibrometry}(a)) with an amplitude of $\approx 3~\mu\mathrm{m}/100\mathrm{V}$. In both cases, the vibration amplitude increased approximately linearly with the input voltage (Fig.~\ref{fig:vibrometry}(b)) and reached $\approx 4~\mu\mathrm{m}$ before saturating due to the Laser Doppler Vibrometer measurement limit, indicating that at higher voltages, the vibration amplitude can be greater. These results confirm that both modules meet the ultrasonic lubrication design requirements, producing amplitudes greater than $2~\mu\mathrm{m}$ at ultrasonic frequencies ($\geq 20~\mathrm{kHz}$).

\begin{figure}
    \centering
    \includegraphics[width=\linewidth]{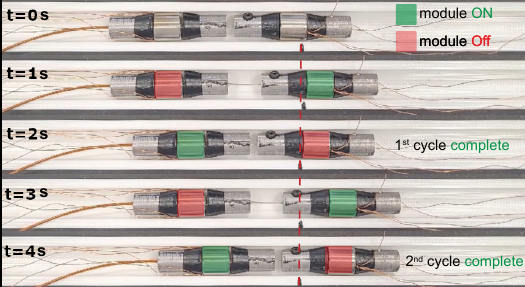}
    \caption{\textbf{Time-lapse sequence of the inchworm locomotion system traversing a rigid PLA track.} By selectively activating ultrasonic lubrication at each friction control module, forward propulsion is achieved following the inchworm motion cycle. Images are shown at one-second intervals. Refer to supplementary videos 1 \& 2 for demonstrations of the use cases presented in this paper, including the inchworm locomotion case.}
    \label{fig:proof-of-concept}
\end{figure}

In the proof-of-concept locomotion experiments, both systems achieved locomotion only when ultrasonic lubrication was activated (Fig.~\ref{fig:proof-of-concept}) and failed to move when it was inactive. Without ultrasonic lubrication, the two friction-control modules experienced symmetric friction and therefore generated no net traction over a cycle. Activating ultrasonic lubrication selectively reduced friction at one module, allowing it to slide while the other remained stationary due to its high friction state. This selective friction control broke the frictional symmetry of the modules, creating the asymmetry required for locomotion. In both systems, bi-directional locomotion was obtained simply by reversing the ultrasonic lubrication activation sequence on the modules (see supplementary videos 1 \& 2). The inchworm-inspired system achieved a mean locomotion efficiency of $94.75\%$, and the ovipositor-inspired system achieved $93.2\%$. These locomotion efficiencies are specific to the experimental setup and normal loading conditions; however, they demonstrate the potential of ultrasonic lubrication as a friction-control mechanism to reliably achieve locomotion and enhance its efficiency.
\begin{figure*}
    \centering
    \includegraphics[width=\linewidth]{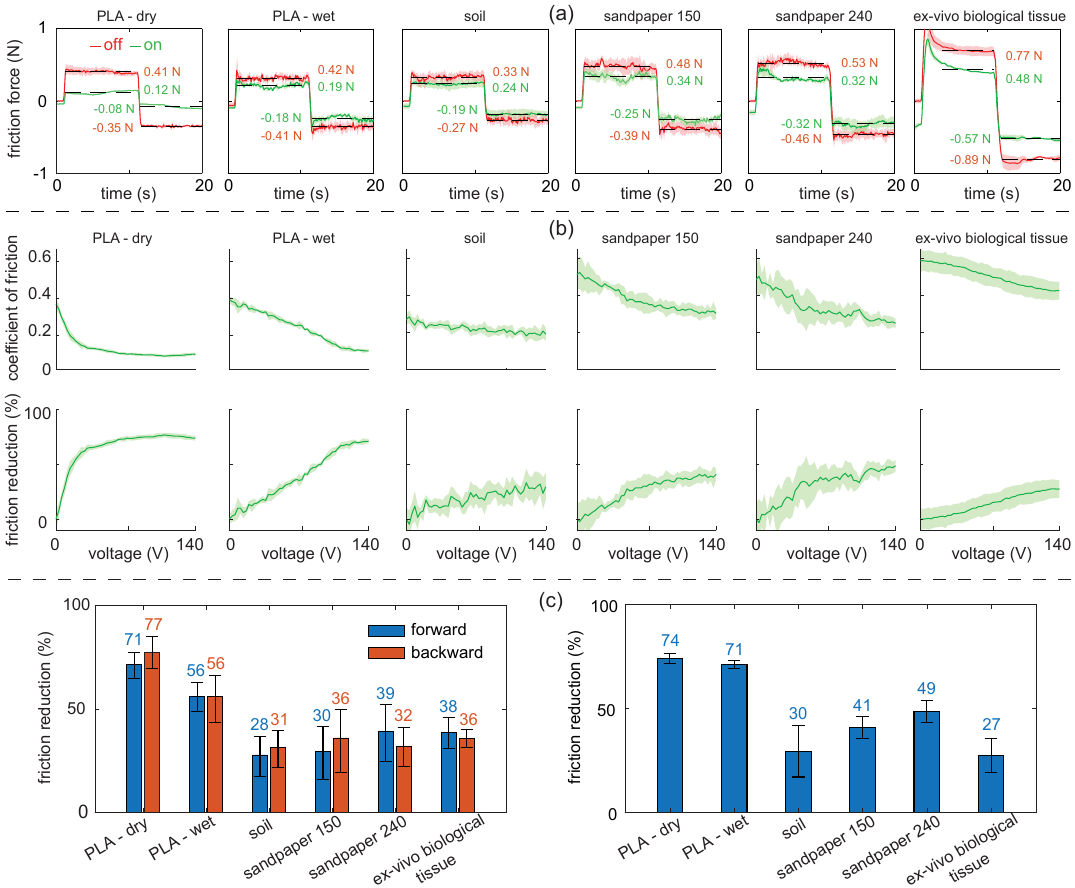}
    \caption{\textbf{Friction modulation experiments across a range of surfaces.} (a) Measured friction forces under forward and backward sliding show that the vibrating slider effectively reduces friction on PLA (dry and wet), soil, sandpaper, and biological tissue (colon). (b) Voltage modulation of the coefficient of friction (top row) and corresponding friction reduction percentage (bottom row). On PLA, reduction is consistent with air squeeze-film formation and exhibits nonlinear saturation. On wet PLA, ultrasonic lubrication enabled friction reduction via liquid squeeze-film formation, following a more linear trend compared to air. Sandpaper surfaces show higher reduction for finer grit (240) compared to coarser grit (150), due to enhanced squeeze-film effects in smaller asperities. Soil produced relatively low friction reduction owing to its unstable granular contact. On colon tissue, viscoelasticity damped vibration energy, decreasing friction reduction. (c) Summary of friction reduction percentage on different substrates during experiments on forward–backward sliding at maximum voltage (left) and experiments with voltage modulation (right).}
    \label{fig:results}
\end{figure*}

The sliding friction experiments demonstrated the viability of ultrasonic lubrication across a range of surfaces (Fig.~\ref{fig:results}), achieving average friction reductions of $71$--$77\%$ on dry PLA, $\approx 56\%$--$71\%$ on wet PLA, $28$--$49\%$ on soil and sandpaper, and $\approx 27$--$38\%$ on \textit{ex-vivo} biological tissue. In the steady-sliding experiments (Fig.~\ref{fig:results}(a)), the flat friction-control module was first slid forward and backward on each surface without activating ultrasonic lubrication to establish the baseline friction forces. The same forward--backward sliding cycles were then repeated with the ultrasonic lubrication activated. In all cases, activating ultrasonic lubrication reduced the measured friction forces compared to the baseline case, and the reduction remained stable over the forward and backward cycles. This shows that ultrasonic lubrication can provide a steady and consistent direction-independent reduction in friction, rather than a transient effect.

The friction modulation experiments (Fig.~\ref{fig:results}(b)) further showed that the ultrasonic lubrication effect is controllable beyond a simple on--off state. In these experiments, the input voltage (and thus the vibration amplitude) was varied to modulate the coefficient of friction. As shown in Fig.~\ref{fig:results}(b), changing the voltage yielded a well-defined, continuous range of friction coefficients across all tested surfaces, demonstrating that ultrasonic lubrication can be used to finely tune friction simply by adjusting the vibration level.

The friction reduction levels and trends across the different substrates help explain the underlying differences in performance, as summarized in Fig.~\ref{fig:results}(c) for both the steady-sliding comparison (left) and the voltage-modulation case (right). On smooth dry PLA, the friction reduction is consistent with the formation of an air squeeze film. Because air is compressible, the film thickness develops nonlinearly with excitation and eventually saturates, which matches the nonlinear trend observed in the voltage modulation experiments (Fig.\ref{fig:results}(b)). On wet PLA, ultrasonic lubrication was still able to reduce friction via the formation of a pressurized liquid squeeze film. Unlike air, liquids are nearly incompressible, and vibrations generate squeeze-film pressure mainly due to liquid inertia~\cite{Atalla2023}, leading to a more linear voltage--friction response compared to air, as shown in Fig.~\ref{fig:results}(b). 

When tested on rough surfaces, the degree of reduction depended on surface roughness. On sandpaper, the finer grit (240) produced slightly greater friction reduction than the coarser grit (150), as the smaller asperities and voids in the finer grit case help promote stronger squeeze-film formation, resulting in an improved lubrication effect. Soil yielded relatively low friction reduction, which can be explained by its granular structure, which creates a constantly changing and non-uniform contact interface, disrupting stable squeeze-film formation, and thus reducing the effect of ultrasonic lubrication. Finally, the \textit{ex-vivo} porcine intestinal tissue interface showed a reduced effect of ultrasonic lubrication compared to that of the rigid wet PLA case, which can be attributed to the viscoelastic properties of tissue. The viscoelasticity of the tissue dissipates part of the vibration energy, which limits the pressure buildup in the squeeze film and thus reduces the lubrication effect.

\section{Conclusion}\label{sec:conclusion}
We introduced ultrasonic lubrication as a friction-control mechanism for robotic locomotion. We presented two friction control module designs: cylindrical and flat, and integrated them into two locomotion systems inspired by inchworm and parasitic wasp ovipositor motion. Vibrometry experiments confirmed that both modules can generate amplitudes exceeding $2~\mu\mathrm{m}$ at ultrasonic frequencies around $22~\mathrm{kHz}$, fulfilling the requirements of ultrasonic lubrication. In proof-of-concept locomotion experiments, ultrasonic lubrication enabled bi-directional locomotion in both inchworm-inspired and ovipositor-inspired systems, with mean efficiencies of $94.75\%$ and $93.2\%$, while no locomotion was achieved when ultrasonic lubrication was not activated due to friction symmetry. Sliding friction experiments demonstrated the ability of ultrasonic lubrication to achieve stable and tunable friction reduction across a variety of rigid and soft interfaces in dry and wet conditions, highlighting the versatility of ultrasonic lubrication for robotic locomotion applications.

In future work, we will focus on modeling the dynamics of ultrasonic lubrication in locomotion, with particular attention to activation timing during the gait cycle. By relating activation phases to locomotion efficiency and stability, we aim to develop optimized friction control strategies for locomotion and explore additional functionalities, such as adaptive steering and programmable interaction with complex environments.

\addtolength{\textheight}{-5.5cm}   




\section*{Acknowledgment}

The authors would like to thank Remi van Starkenburg, from the Central Workshop of TU Delft (DEMO), for support in fabricating the modules and Maurits Pfaff, Cognitive Robotics department of TU Delft, for support in integrating the piezoelectric plates.


\bibliographystyle{ieeetr}
\bibliography{biblography}



\end{document}